\documentclass[10pt,twocolumn,letterpaper]{article}

\usepackage[pagenumbers]{cvpr} 

\definecolor{cvprblue}{rgb}{0.21,0.49,0.74}
\usepackage[pagebackref,breaklinks,colorlinks,allcolors=cvprblue]{hyperref}

\usepackage{pifont,textcomp}
\usepackage{algorithm}
\usepackage{algorithmic}
\usepackage{amsmath}
\usepackage{amssymb}

\usepackage{amsmath,amsfonts,bm,amsthm}

\def\eqref#1{equation~\ref{#1}}

\def\1{\bm{1}}

\def\rvepsilon{{\mathbf{\epsilon}}}
\def\rveps{{\rvepsilon}}

\def\rvx{{\mathbf{x}}}

\def\vu{{\bm{u}}}
\def\vv{{\bm{v}}}

\def\vx{{\bm{x}}}

\def\mI{{\bm{I}}}

\DeclareMathAlphabet{\mathsfit}{\encodingdefault}{\sfdefault}{m}{sl}
\SetMathAlphabet{\mathsfit}{bold}{\encodingdefault}{\sfdefault}{bx}{n}

\newcommand{\bx}{{\bf x}}

\newcommand{\noisyx}{\tilde{\bx}}

\newcommand{\esm}{\mathcal{L}_\text{ESM}}

\newcommand{\esmglobal}{\mathbb{E}_{\noisyx} \left[ \esm(\noisyx,\, \sigma,\, \theta) \right]}

\newcommand{\dsmglobal}{\mathbb{E}_\rvx\left[\mathcal{L}_\text{DSM}(\rvx,\, \sigma,\, \theta)\right]}

\newcommand{\gcheck}{\textcolor{green}{\ding{52}}}
\newcommand{\rxmark}{\textcolor{red}{\ding{56}}}

\newtheorem{theorem}{Theorem}[section]  

\newtheorem*{theorem*}{Theorem}
\newtheorem*{lemma*}{Lemma}

\theoremstyle{definition}
\newtheorem{definition}[theorem]{Definition}

\theoremstyle{remark}

\def\paperID{29773} 
\def\confName{CVPR}
\def\confYear{2026}

\title{Saddle-Free Guidance:\\Improved On-Manifold Sampling without Labels or Additional Training}

\author{Eric Yeats \quad Darryl Hannan \quad Wilson Fearn \quad Tim Doster \quad Henry Kvinge \quad Scott Mahan \\
Pacific Northwest National Laboratory\\
Seattle, Washington USA\\
{\tt\small \{first\}.\{last\}@pnnl.gov}
}

\begin{document}
\maketitle
\begin{abstract}
Score-based generative models require guidance in order to generate plausible, on-manifold samples. The most popular guidance method, Classifier-Free Guidance (CFG), is only applicable in settings with labeled data and requires training an additional unconditional score-based model. More recently, Auto-Guidance adopts a smaller, less capable version of the original model to guide generation. While each method effectively promotes the fidelity of generated data, each requires labeled data or the training of additional models, making it challenging to guide score-based models when (labeled) training data are not available or training new models is not feasible.

We make the surprising discovery that the positive curvature of log density estimates in saddle regions provides strong guidance for score-based models. Motivated by this, we develop saddle-free guidance (SFG) which maintains estimates of maximal positive curvature of the log density to guide individual score-based models. SFG has the same computational cost of classifier-free guidance, does not require additional training, and works with off-the-shelf diffusion and flow matching models. Our experiments indicate that SFG achieves state-of-the-art FID and FD-DINOv2 metrics in single-model unconditional ImageNet-512 generation. When SFG is combined with Auto-Guidance, its unconditional samples achieve general state-of-the-art in FD-DINOv2 score. Our experiments with FLUX.1-dev and Stable Diffusion v3.5 indicate that SFG boosts the diversity of output images compared to CFG while maintaining excellent prompt adherence and image fidelity.
\end{abstract}    
\begin{figure}[t!]
    \centering
    \begin{subfigure}{0.95\linewidth}
        \centering
        \includegraphics[width=0.99\linewidth,trim=5 0 0 0, clip]{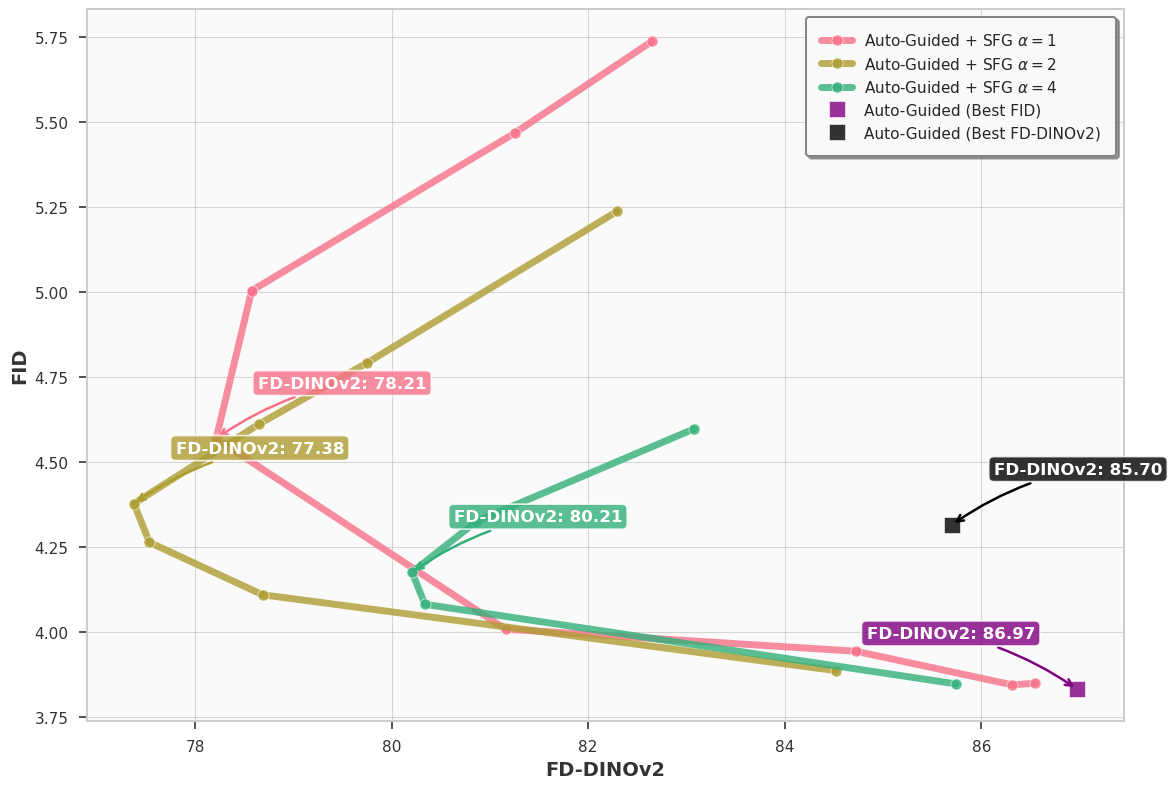}
        \caption{SFG boosts SOTA FD-DINOv2 scores on unconditional ImageNet.}\label{subfig:ag_sfg_uncond}
    \end{subfigure}

    \begin{subfigure}{0.24\linewidth}
        \includegraphics[width=0.99\linewidth]{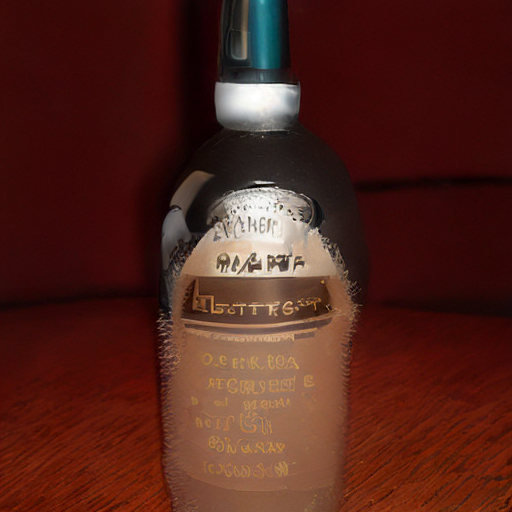}
    \end{subfigure}
    \hfill
    \begin{subfigure}{0.24\linewidth}
        \includegraphics[width=0.99\linewidth]{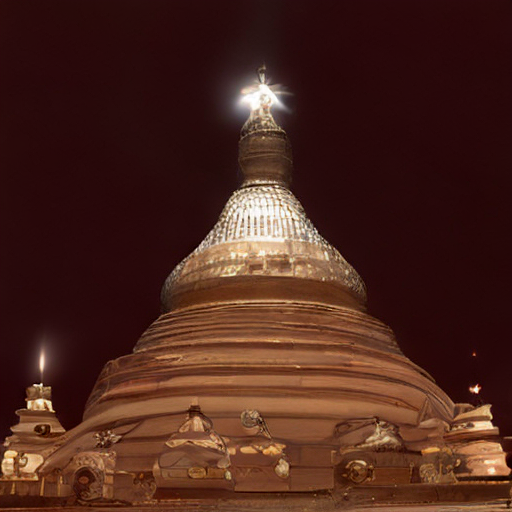}
    \end{subfigure}
    \hfill
    \begin{subfigure}{0.24\linewidth}
        \includegraphics[width=0.99\linewidth]{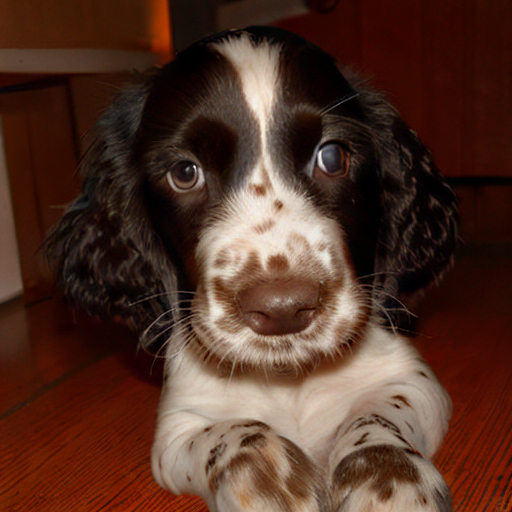}
    \end{subfigure}
    \hfill
    \begin{subfigure}{0.24\linewidth}
        \includegraphics[width=0.99\linewidth]{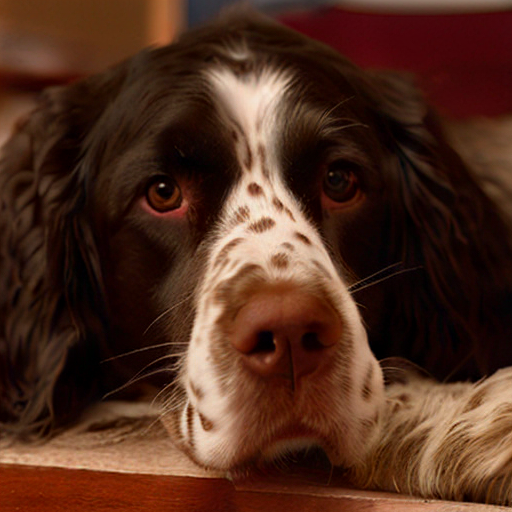}
    \end{subfigure}


    \begin{subfigure}{0.24\linewidth}
        \includegraphics[width=0.99\linewidth]{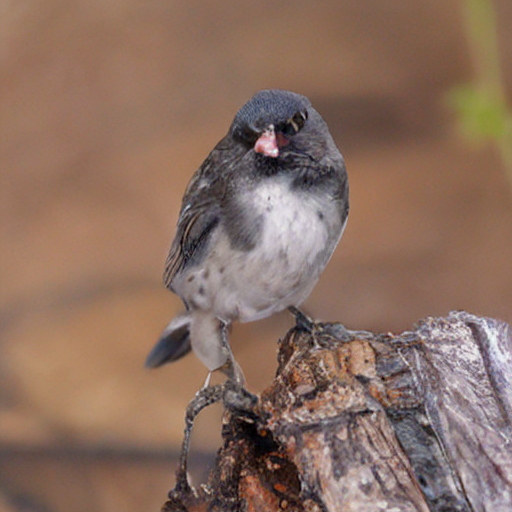}
    \end{subfigure}
    \hfill
    \begin{subfigure}{0.24\linewidth}
        \includegraphics[width=0.99\linewidth]{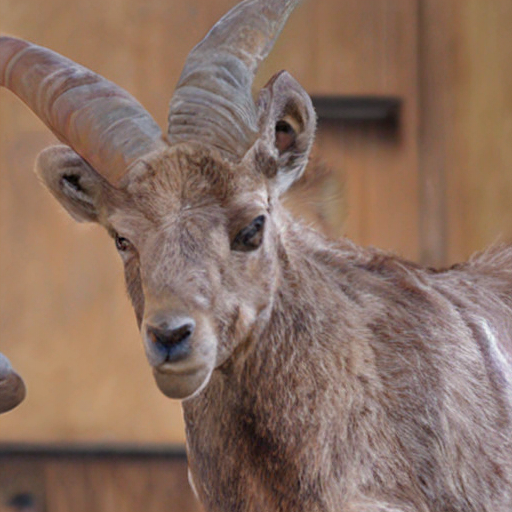}
    \end{subfigure}
    \hfill
    \begin{subfigure}{0.24\linewidth}
        \includegraphics[width=0.99\linewidth]{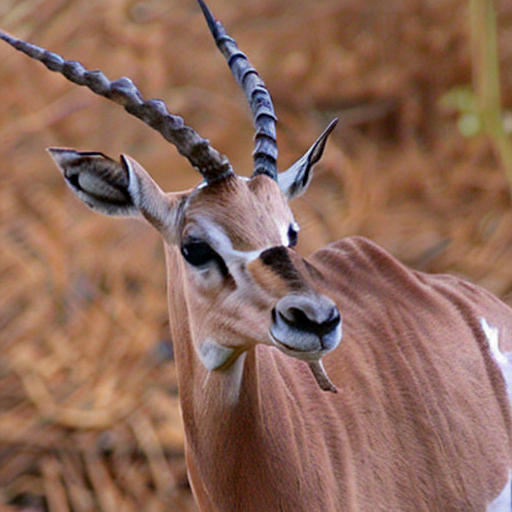}
    \end{subfigure}
    \hfill
    \begin{subfigure}{0.24\linewidth}
        \includegraphics[width=0.99\linewidth]{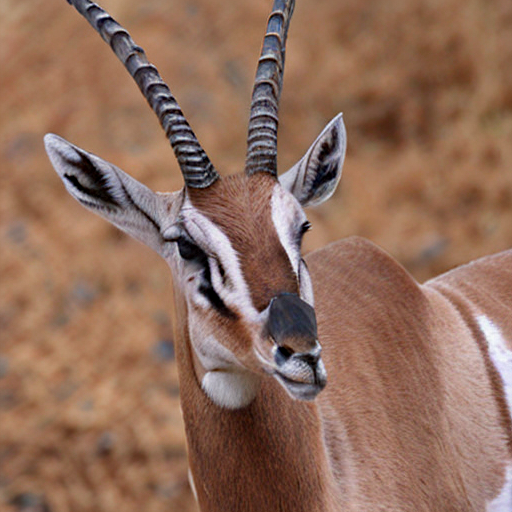}
    \end{subfigure}

    \begin{subfigure}{0.24\linewidth}
        \includegraphics[width=0.99\linewidth]{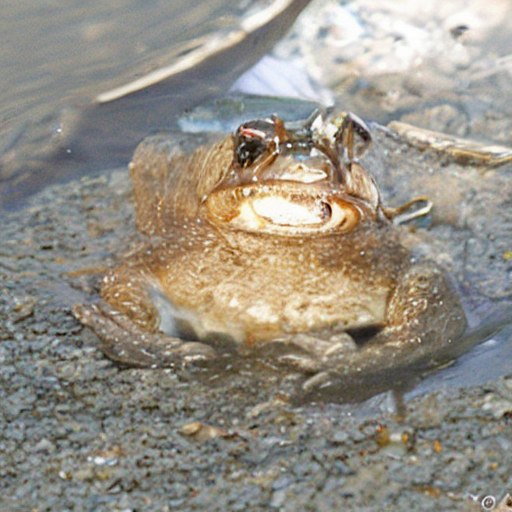}
        \caption{EDM2 Baseline}
    \end{subfigure}
    \hfill
    \begin{subfigure}{0.24\linewidth}
        \includegraphics[width=0.99\linewidth]{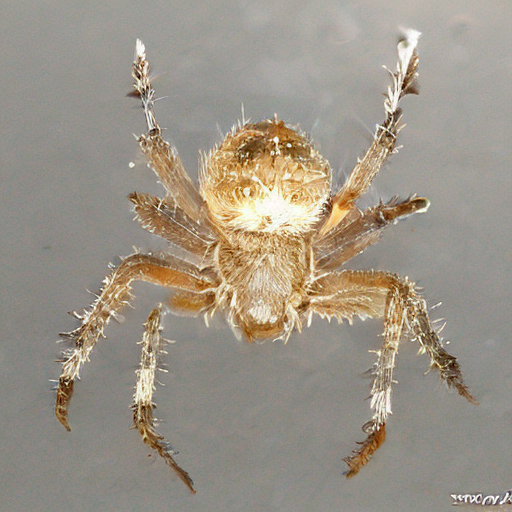}
        \caption{EDM2 + SFG (Ours)}
    \end{subfigure}
    \hfill
    \begin{subfigure}{0.24\linewidth}
        \includegraphics[width=0.99\linewidth]{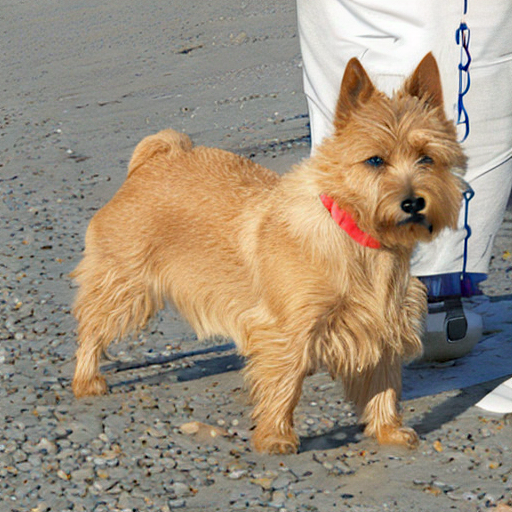}
        \caption{EDM2 + Auto Guidance (AG)}
    \end{subfigure}
    \hfill
    \begin{subfigure}{0.24\linewidth}
        \includegraphics[width=0.99\linewidth]{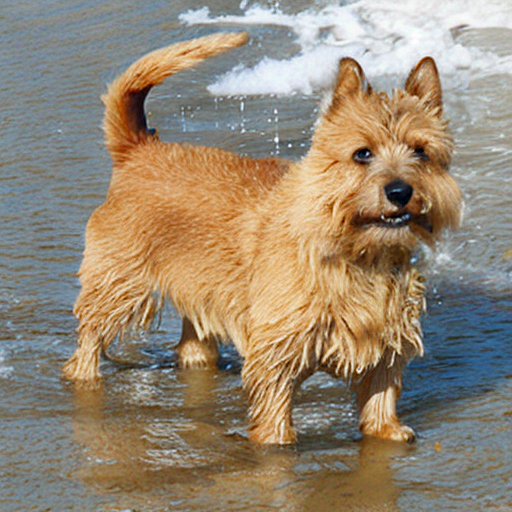}
        \caption{EDM2 + AG + SFG (Ours)}
    \end{subfigure}
    \caption{SFG improves image realism by steering samples away from saddle regions. SFG combined with Auto-Guidance achieves state-of-the-art FD-DINOv2 on unconditional ImageNet-512.}
\end{figure}

\section{Introduction}

In recent years, diffusion and flow matching models have dethroned generative adversarial networks as the preferred generative model in many applications \cite{dhariwal2021diffusion,stein2023exposing}. Broadly referred to as score-based generative models \cite{song2020score}, diffusion \cite{ho2020denoising} and flow matching \cite{lipman2022flow} models dominate in the fields of image generation \cite{esser2024scaling,karras2022elucidating}, video generation \cite{blattmann2023stable,yang2024cogvideox,xing2024survey}, audio generation \cite{liu2023audioldm,huang2023make}, media editing \cite{yeats2024counterfactual,huang2025diffusion}, out-of-distribution detection \cite{liu2023unsupervised,graham2023denoising}, and inverse problems \cite{song2020score,moliner2023solving}, to name a few. 

One of the key ingredients to high quality generative modeling is the ``truncation trick''. Originally developed as a simple noise resampling method for generative adversarial networks (GANs) \cite{brock2018large}, the truncation trick allows one to trade off sample diversity for sample fidelity. The analogous method in score-based models is broadly referred to as \textit{guidance} and was first studied by \citet{dhariwal2021diffusion}. They observed that injecting stronger levels of classifier gradient into the conditional diffusion model sampling process significantly improved sample fidelity. \citet{ho2022classifier} proposed ``Classifier-Free Guidance'' (CFG). In the CFG method, they proposed that the classifier be dropped entirely and that an unconditional diffusion model be amortized into the original model to provide a guiding signal during generation. CFG has since become the dominant guidance method, but it is only applicable to conditional sampling. As a result, the performance of unconditional sampling has lagged far behind that of conditional sampling \cite{karras2022elucidating}.

A growing number of works have started to address the unconditional guidance problem. \citet{ahn2024self} introduce ``Perturbed-Attention Guidance'' (PAG), an inference-time method which creates guiding signals by replacing the self-attention patterns of diffusion U-Nets \cite{ronneberger2015u} with identity matrices. Similarly, \citet{hong2024smoothed} introduces ``Smoothed Energy Guidance'' (SEG), another inference-time method which generates guiding signals by blurring the self-attention maps of diffusion U-Nets with 2D Gaussian convolutions. A landmark paper by \citet{karras2024guiding} proposes Auto-Guidance (AG), where sampling is guided by a less capable version of the original model. Like CFG, AG requires training an additional guiding network. But, unlike CFG, AG is not restricted to settings with labels, and it achieves state-of-the-art scores in both conditional and unconditional generation on ImageNet \cite{russakovsky2015imagenet}.

While there has been much progress in guidance research, each method either requires labels \cite{ho2022classifier,kynkaanniemi2024applying}, requires additional training \cite{karras2024guiding}, is limited to specific architectures \cite{ahn2024self,hong2024smoothed}, or is limited by a combination thereof. The crux of the shared limitations of current methods is that they each guide sampling of the original distribution with a different, more entropic reference distribution, whether it is specifically trained for or simulated through model interventions.

We challenge the current guidance paradigm and hypothesize that sampling with score-based models is hindered specifically by \textit{saddle regions} of the learned log density. We corroborate this hypothesis in a controlled experiment and make the surprising discovery that the defining characteristic of saddle regions, positive curvature, provides a strong guidance signal to score-based models. We propose Saddle-Free Guidance (SFG), a model-agnostic, training- and label-free guidance method based purely on the geometry of the modeled log density. Our work provides the following contributions:
\begin{itemize}
    \item Saddle-Free Guidance (SFG), a novel inference-time guidance strategy which relies on the geometry of the modeled log density to boost the fidelity of samples while maintaining distributional diversity
    \item A comparison of SFG with training-time and inference-time guidance methods on the large scale ImageNet 512x512 benchmark and the foundational text to image models Stable Diffusion 3.5 and Flux.1-dev
    \item State-of-the-art single-model unconditional generation performance on ImageNet512x512 and general state-of-the-art unconditional generation performance on ImageNet512x512 when combined with Auto-Guidance \cite{karras2024guiding}
\end{itemize}

\section{Background and Related Work}

\paragraph{Score-based Models} Let us consider modeling of the data distribution $p(\rvx)$ which generates samples $\rvx \in \mathbb{R}^n$. Score-based models $s_\theta(\cdot): \mathbb{R}^n \rightarrow \mathbb{R}^n$, parameterized by $\theta$, capture the data distribution $p(\rvx)$ by approximating the \textit{score} $\nabla \log p(\vx)$. For data $\noisyx \sim (g_\sigma * p)(\noisyx)$ where $g_\sigma$ is a Gaussian density of scale $\sigma$, $s_\theta(\cdot)$ should ideally minimize:
\begin{multline}\label{eqn:esm_loss}
    \esmglobal = \\
    \mathbb{E}_{\noisyx \sim p(\noisyx)} \left[ \sigma^2\,\left\| \nabla\log p(\noisyx) - s_\theta(\noisyx) \right\|^2 \right].
\end{multline}
However, the ground truth score $\nabla \log p(\noisyx)$ is generally intractable and unknown, making this \textit{explicit} score matching objective challenging to use in practice. \citet{vincent2011connection} proved that the tractable objective \textit{denoising} score matching $\dsmglobal$ is an equivalent objective to \cref{eqn:esm_loss}: 
\begin{multline}\label{eqn:dsm_loss}
    \dsmglobal = \\
    \mathbb{E}_{\rvx \sim p(\rvx), \rveps \sim p(\rveps)} \left[ \left\| \rveps - \rveps_\theta(\rvx + \sigma\rveps) \right\|^2 \right],
\end{multline}
where $\rveps_\theta(\rvx)=-\sigma s_\theta(\rvx)$ is a noise prediction parameterization of the score model \cite{ho2020denoising}. Generally speaking, diffusion and Gaussian flow models are sequences of score models trained on varying scales of noise $\sigma$ which bridge the data distribution to a simple Gaussian distribution (we omit the dependence of the score models on $\sigma$ for simplicity). Sampling involves integrating the score estimates of the sequence of score models using ODE solvers \cite{lipman2022flow}, SDE solvers \cite{ho2020denoising,song2020score}, or Langevin dynamics \cite{song2019generative}.

\paragraph{Guidance} \citet{dhariwal2021diffusion} observed that augmenting score estimates with the gradient of a classifier $f_\phi(\cdot): \mathbb{R}^n \rightarrow [0, 1]^k$, where $\phi$ is the set of classifier parameters and $k$ is the class count, significantly improved sample quality. They weighed this gradient with $w_{\text{DN}}$, yielding the guided score estimate:
\begin{equation}
    s^{\text{DN}}_{\theta,\phi}(\vx, y_i) = s_\theta(\vx) + w_{\text{DN}} \nabla\log f_\phi(\vx)_i,
\end{equation}
where $i$ indexes the class output. \citet{ho2022classifier} observed that $\nabla \log f_\phi(\vx)_i \approx \nabla \log p(\vx|y=i) - \nabla \log p(\vx)$, and they proposed Classifier-Free Guidance (CFG). The main idea is to guide generation with the difference between a conditional score model $s_\theta(\vx, y)$ with an unconditional score model $s_\phi(\vx)$, yielding the guided score estimate:
\begin{equation}
    s^{\text{CFG}}_{\theta,\phi}(\vx, y_i) = s_\theta(\vx, y_i) + (w_{\text{CFG}}-1) \left( s_\theta(\vx, y_i) - s_\phi(\vx)\right),
\end{equation}
where $w_{\text{CFG}} \geq 1$ controls the strength of the guidance. CFG is very effective in enhancing the fidelity of generated samples, however it requires labelled data, it requires training an additional model, and it can significantly reduce sample diversity.

Unconditional (U) guidance methods such as Auto-Guidance (AG) \cite{karras2024guiding}, Perturbed-Attention Guidance (PAG) \cite{ahn2024self}, and Smoothed Energy Guidance (SEG) \cite{hong2024smoothed} all follow a similar form inspired by CFG:
\begin{equation}
    s^{\text{U}}_{\theta,\phi}(\vx) = s_\theta(\vx) + (w_{\text{U}}-1) \left( s_\theta(\vx) - s_\phi(\vx) \right),
\end{equation}
where $w_{\text{U}} \geq 1$ and $s_\phi(\cdot)$ is a `bad', more entropic version of the original score model $s_\theta(\cdot)$. $s_\phi(\cdot)$ can be realized by training a smaller model with less data (AG) or simulated by intervening on the parameters or computation of the original model (PAG and SEG). Unconditional guidance methods may be applied to conditional sampling by simply using conditional $s_\theta(\vx, y)$ and $s_\phi(\vx, y)$.

\subsection*{Score-Based Models and Saddle Regions}\label{sec:saddle_simplex}

We design a simple experiment to test the hypothesis that saddle regions of the log data density pose specific challenges to score-based models. We use flow matching \cite{lipman2022flow} to train a series of diffusion transformer models \cite{peebles2023scalable} of increasing parameter and flop count on a simplex Gaussian mixture model (GMM) task. The simplex GMM consists of 16 Gaussian components of scale $0.2$ immersed in a 256-dimensional space, where each Gaussian component is situated at a vertex of a 16-simplex. This GMM structure was chosen to ensure that each Gaussian component is a fixed distance of $\sqrt{2}$ from all other Gaussian components, providing clear mode, saddle, and outlier regions in a nontrivial setting. We characterized `mode' data as that sampled from the original 16 Gaussian components, `saddle' data as that sampled from similar Gaussian components situated directly in-between any pair of the original $16$ components (bisecting each edge), and `outlier' data as that sampled from a similar, twice as large 16-simplex that encompasses the original 16-simplex.

\begin{figure}[ht]
    \centering
    \includegraphics[width=0.8\linewidth,trim=5 0 0 0, clip]{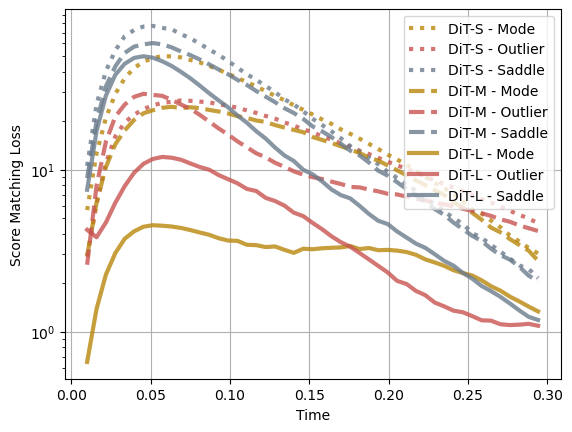}
    \caption{Explicit score matching loss (\ref{eqn:esm_loss}) on the test set for the $16$-simplex task. Test points within saddle regions experience the highest loss and benefit the least from DiT scaling.}\label{fig:simplex_esm}
\end{figure}

\cref{fig:simplex_esm} depicts the explicit score matching loss of the trained DiT models on mode, saddle, and outlier test data at various flow matching times. As expected, increasing parameter and FLOP count (from DiT-S to DiT-L) tends to decrease the explicit score matching loss of each rectified flow. However, the explicit score matching loss is larger for saddle data rather than that for mode or outlier data across most time scales. Moreover, the explicit score matching loss for saddle data benefits the \textit{least} from transformer scaling. From the DiT-S to the DiT-L, the explicit score matching loss decreases by a factor of ~10 for mode data and a factor of ~3 for outlier data, but it only decreases by a factor of ~1.3 for saddle data. These results support the hypothesis that score-based models fit saddle regions relatively poorly, and that scaling models is not an effective means of improving score matching performance in saddle regions.
\section{Saddle-Free Guidance (SFG)}

\begin{table*}[t!]
  \caption{High-level comparison of Saddle-Free Guidance (SFG) with popular guidance strategies.}
  \label{tab:qual_comparison}
  \centering
  \begin{tabular}{@{}lcccccc@{}}
    \toprule
    Desideratum & CFG \cite{ho2022classifier} & CFG (Interval) \cite{kynkaanniemi2024applying} & Auto-Guidance \cite{karras2024guiding} & SEG \cite{hong2024smoothed} & PAG \cite{ahn2024self} & SFG (Ours) \\
    \midrule
    Samples are On-Manifold & \gcheck & \gcheck & \gcheck & \gcheck & \gcheck & \gcheck \\
    Preserves Sample Diversity & \rxmark & \gcheck & \gcheck & \gcheck & \gcheck & \gcheck \\
    Unconditional Sampling & \rxmark & \rxmark & \gcheck & \gcheck & \gcheck & \gcheck \\
    Only Requires One Model & \rxmark & \rxmark & \rxmark & \gcheck & \gcheck & \gcheck \\
    Model-Agnostic & \gcheck & \gcheck & \gcheck & \rxmark & \rxmark & \gcheck \\
    \bottomrule
  \end{tabular}
\end{table*}

We hypothesized that sample quality from diffusion and flow matching models is hampered by poor score estimates in saddle regions. This was corroborated by our simplex experiment from \cref{sec:saddle_simplex}. Next, we develop a method to identify points which belong to saddle regions of the log density, and we show that this naturally leads to an effective guidance strategy.

\begin{definition}[Saddle Points]
The set of saddle points of the log density $\log p(\rvx)$ is:
\begin{equation}
    \mathcal{S} = \left\{\vx : \nabla \log p(\vx) = 0 \text{ and } \nabla^2 \log p(\vx) \text{ is indefinite}\right\},
\end{equation}
where $\nabla^2 \log p(\vx)$ is the Hessian of the log density at $\vx$.
\end{definition}
In practice, the score $\nabla \log p(\vx)$ is approximated with a trained score network $s_\theta(\vx)$, yielding the approximate set:
\begin{equation}\label{eqn:approx_saddle_def}
    \mathcal{S} \approx \left\{\vx : s_\theta(\vx) = 0 \text{ and } \nabla s_\theta(\vx) \text{ is indefinite}\right\},
\end{equation}
where $\nabla s_\theta(\vx)$ is the Jacobian of the score network. While \cref{eqn:approx_saddle_def} can identify saddle points of the approximate log density, it is not useful in practical scenarios since $s_\theta(\vx)$ is almost never zero, and computing $\nabla s_\theta(x)$ is computationally demanding. Instead, we relax this condition to define \textit{saddle regions} set $\mathcal{S}_{\text{SFG}}$ as:
\begin{equation}
    \mathcal{S}_{\text{SFG}} = \left\{\vx : \sup \left(\, \text{Sp}\left(\nabla s_\theta(\vx)\right) \,\right)\ >\ 0\right\},
\end{equation}
where $\text{Sp}(\cdot)$ is the matrix spectrum. In other words, points $\vx$ are considered to belong to a saddle region if the maximal eigenvalue of the score network Jacobian at $\vx$ is positive. This formulation is more pragmatic in that it encompasses a much larger set of points, and the most positive eigenvalue $\lambda_+$ of the score network Jacobian can be computed feasibly by solving:
\begin{equation}
    \lambda_+ = \sup_\vv\left\{ \left\|\left(\nabla s_\theta(\vx) + \alpha \mI\right) \vv \right\| - \alpha : \vv \in \mathbb{R}^n\ \text{s.t.}\ \|\vv\| \leq 1\right\},
\end{equation}
where the shift constant $\alpha>0$ ensures all eigenvalues of the matrix $\left(\nabla s_\theta(x) + \alpha I\right)$ are positive. Crucially, the supremum can be computed efficiently using \textit{shifted power iteration}\footnote{If the shift is not applied, power iteration could converge to the eigenvector of the most \textit{negative} eigenvalue if its magnitude is largest.}, and the inner Jacobian-vector product can be approximated without explicit gradient computation:
\begin{multline}
    \left(\nabla s_\theta(\vx) + \alpha \mI\right) \vv \approx \vu, \\ \vu := \frac{s_\theta(\vx + h\vv) - s_\theta(\vx)}{h\, \|\vv\|} + \alpha \vv,
\end{multline}
where $h \rightarrow 0^+$ is a small positive constant. In summary, power iteration of $(\nabla s_\theta(\vx) + \alpha\mI)$ yields vectors $\vv$ and $\vu$, where $\vv$ is an eigenvector of $\nabla s_\theta(\vx)$, $\vu \approx \nabla s_\theta(\vx) \vv + \alpha\vv$, and thus $\lambda_+ = \vu^T\vv - \alpha$. 

Our insight is that in saddle regions (i.e., $\lambda_+ > 0$), the maximal log density curvature vector $\vu - \alpha\vv$ may be added to the score estimate to steer generation away from saddle regions. Concretely, we augment score estimates with:
\begin{equation}
    s^{\text{SFG}}_\theta(\vx) = s_\theta(\vx) + w_{\text{SFG}}\, \sigma \, H(\lambda_+) (\vu - \alpha \vv),
\end{equation}
where $w_{\text{SFG}} \geq 0$ and $H(\cdot)$ is the Heaviside step function. Score networks are often parameterized as $\rveps_\theta(\vx)=-\sigma s_\theta(\vx)$, where $\rveps_\theta(\cdot)$ is a ``noise prediction function'' and $\sigma$ is the scale of the noise added during training. In this case, the update is:
\begin{equation}
    \rveps^{\text{SFG}}_\theta(\rvx) = \rveps_\theta(\vx) -  w_{\text{SFG}}\, \sigma^2 H(\lambda_+) (\vu - \alpha \vv).
\end{equation}
Since the most negative eigenvalue of $\sigma^2\nabla^2 \log p(\vx) \geq -1$ \cite{yeats2023adversarial,yeats2025connection}, we may simply set $\alpha \geq 1$ to ensure that power iteration converges to the most positive eigenvalue.

\subsection*{Practical Implementation of SFG}

Saddle-Free Guidance (SFG) requires computing power iteration to estimate $\lambda_+$ at each sampling step. Since power iteration often requires multiple iterations to converge, this would be prohibitively expensive in the multi-step sampling process where one would compute $\vv_i$ and $\vu_i$ at each step $i$.

To circumvent this computational cost, we initialize $\vv_{i} = \vu_{i-1}/\|\vu_{i-1}\|$, where $\vu_{i-1}$ is $\vu$ computed from the previous sampling step\footnote{$\vv_0$ is initialized as a uniformly random unit vector.}. This allows the power iteration to converge with much fewer steps, and we use just one step of power iteration in our experiments. This leads to a computational cost which is effectively the same as classifier-free guidance.

Lastly, we dynamically adjust the shift parameter $\alpha$ to ensure that $\lambda_+$ converges to the maximal eigenvalue. Please refer to \cref{alg:sfg} for the full details.

\begin{algorithm}[h]
\caption{Saddle-Free Guidance}\label{alg:sfg}
\begin{algorithmic}[1]
    \REQUIRE Function $\rveps_\theta: \mathbb{R}^n \to \mathbb{R}^n$, input point $\vx \in \mathbb{R}^n$, perturbation vector $\vv \in \mathbb{R}^n$ with $\|\vv\| = 1$, finite difference step $h > 0$, noise scale $\sigma > 0$, shift param. $\alpha \geq 0$, saddle-free guidance weight $w_{\text{SFG}} \geq 0$
    \STATE $\hat{\epsilon} = \rveps_\theta(\vx)$ \hfill\# initial noise estimate
    \STATE $\vu = \frac{\hat{\epsilon} - \rveps_\theta(\vx + h \sigma \vv)}{h}$ \hfill \# $\vu \approx -\sigma \frac{\partial \rveps_\theta(\vx)}{\partial \vx} \vv$
    \STATE $\lambda_+ = \vu^T \vv$ \hfill\# estimate of most positive eigenvalue
    \STATE $\alpha = \max(\alpha,\ -\lambda_+)$ \hfill\# update shift if necessary
    \STATE $m = H(\lambda_+)$ \hfill\# positive curvature only
    \STATE $\hat{\epsilon} = \hat{\epsilon} - m\,w_{\text{SFG}}\,\vu$ 
    \hfill \# apply SFG guidance
    \STATE $\vu = \vu + \alpha\, \sigma\, \vv$ \hfill\# add shift for next iteration
    \STATE $\vv = \vu\, /\, \|\vu\|$ \hfill\# make unit $\vv$ for next iteration
    
    \RETURN $\hat{\epsilon}$, $\vv$, $\alpha$ \hfill\# next noise pred., perturbation, shift
\end{algorithmic}
\end{algorithm}

\section{Experiments}

\begin{figure}[ht]
    \centering
    \includegraphics[width=0.99\linewidth,trim=100 0 100 0, clip]{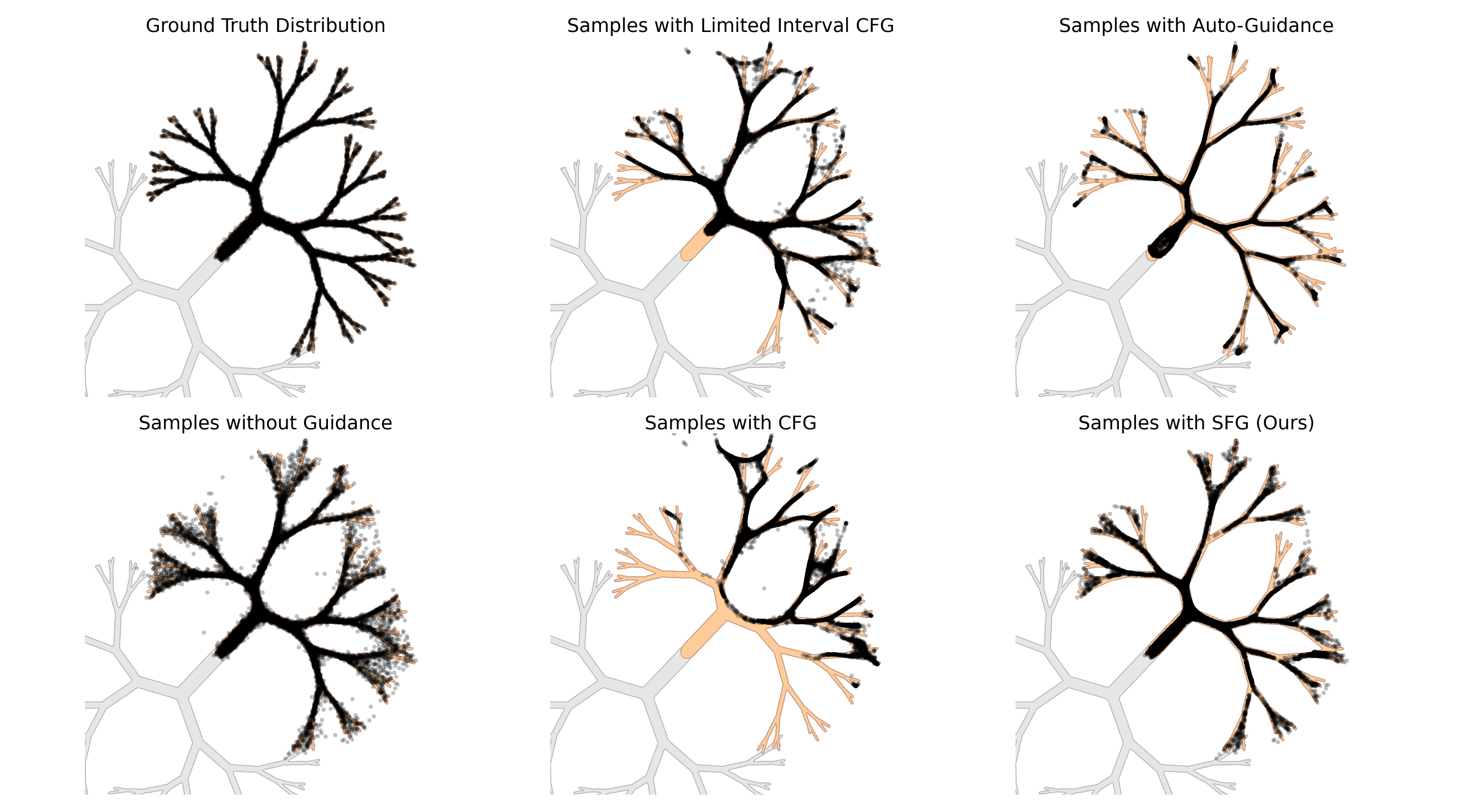}
    \caption{Comparison of sampling without guidance, with classifier-free guidance, with auto-guidance, and with saddle-free guidance on the toy fractal manifold learning task from \citet{karras2024guiding}. Sampling without guidance produces many unlikely `outliers' which are caught in saddle regions between modes. CFG and its variants reduce outlier in saddle regions, but they bias generation and reduce sample diversity. Auto-guidance avoids outliers in saddle regions, but it is overly restrictive and reduces diversity on the manifold. Saddle-free guidance prevents outliers while maintaining sample diversity on the manifold.}\label{fig:fractal_manifold}
\end{figure}

We divide the guidance techniques considered in this work into two groups which (1) leverage additional trained networks or (2) rely on the original network only. Group (1) consists of Classifier-Free Guidance (CFG) \cite{ho2022classifier}, Limited Interval CFG \cite{kynkaanniemi2024applying}, and Auto-Guidance \cite{karras2024guiding}. Group (2) consists of Smoothed Energy Guidance (SEG) \cite{hong2024smoothed}, Perturbed Attention Guidance (PAG) \cite{ahn2024self}, and our method Saddle-Free Guidance (SFG). For a high-level comparison of different guidance methods, see \cref{tab:qual_comparison}.

The effect of the model-agnostic guidance methods can be visualized in \cref{fig:fractal_manifold} with a toy fractal manifold learning problem adapted from \citet{karras2024guiding}. Each subfigure of \cref{fig:fractal_manifold} depicts the conditional samples obtained from sampling the same model without guidance, with an unconditional model (CFG), with an unconditional model in a limited interval (Limited Interval CFG), with a less capable version of the model (Auto-Guidance), and with the original model's positive curvature (SFG), respectively. Sampling without guidance (bottom left) leads to many outliers which get caught in saddles between the branches of the fractal. Sampling with CFG or Limited Interval CFG (center) reduces outliers but biases generation and reduces diversity. Sampling with Auto-Guidance (top right) eliminates outliers but restricts diversity on the manifold. Saddle-Free Guidance (bottom left) eliminates outliers and maintains diversity on the manifold without the need for a new model.

\subsection{ImageNet-512 Experiments}

We compare the various guidance methods on an ImageNet \cite{russakovsky2015imagenet} generation task. Starting from the same model and random seed, we apply a guidance method to sample 50k 512x512 images for downstream comparison with ground truth ImageNet 512x512 images. We leverage PyTorch \cite{paszke2019pytorch} with 8 H100 NVIDIA GPUs. We adapt the sampling and evaluation code from \citet{karras2024guiding} to conduct our experiments.

\paragraph{Metrics} We employ the classical Fréchet Inception Distance (FID) \cite{heusel2017gans} and more recent Fréchet DINOv2 Distance (FD-DINOv2) \cite{stein2023exposing} between generated images and ground truth ImageNet images as our image quality metrics. In each case, lower scores are better, indicating that the generated images are harder to distinguish from the ground truth distribution. \citet{stein2023exposing} showed in a large-scale study that FD-DINOv2 correlates more strongly with the rate at which humans misclassify the generated images as `real' rather than `fake', so we consider FD-DINOv2 as the primary image fidelity metric.


\paragraph{Models} We employ the EDM2 family of diffusion models \cite{karras2024analyzing} for our quantitative experiments. EDM2 models achieve state-of-the-art performance in ImageNet-512 generation and are publicly available for standard comparison of guidance techniques. Moreover, EDM2 models are U-Nets \cite{ronneberger2015u} with self-attention layers, and therefore are compatible with SEG and PAG.

\begin{figure}[ht]
    \centering
    \includegraphics[width=0.8\linewidth,trim=5 0 0 0, clip]{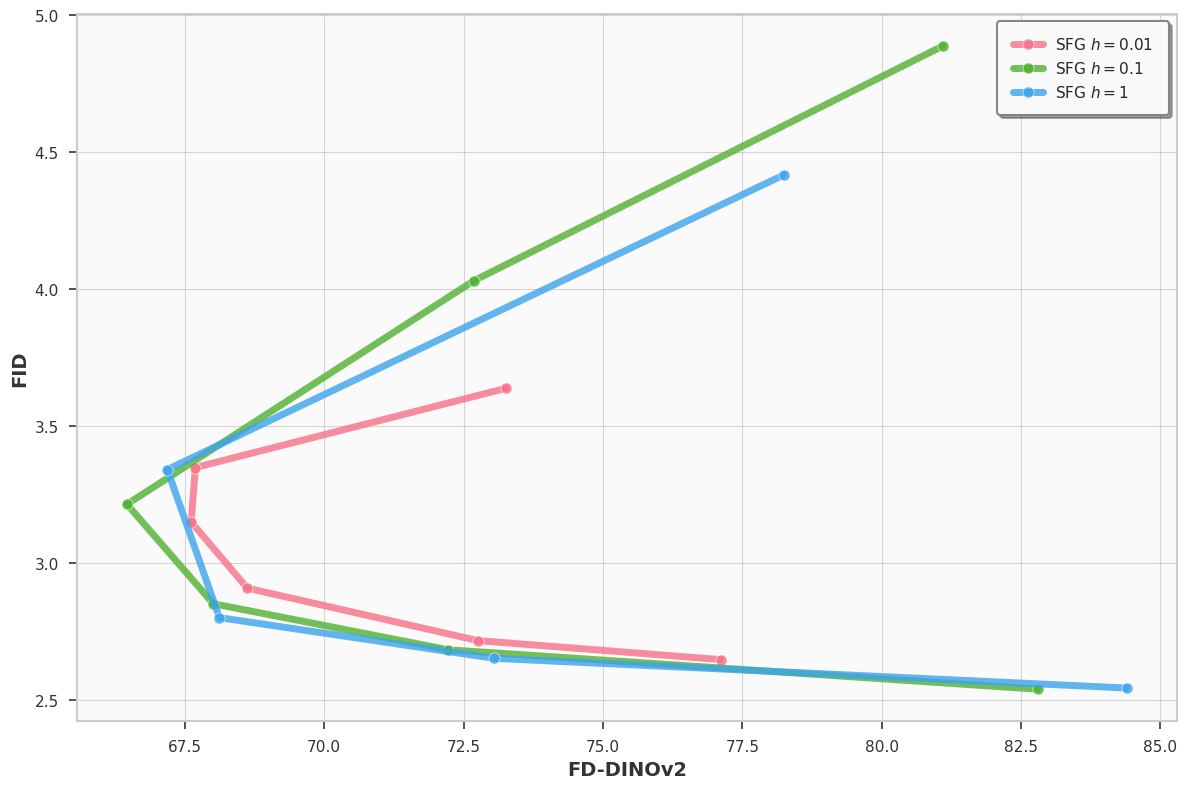}
    \caption{Effect of step size parameter on the FID and FD-DINOv2 tradeoff for EDM2-s-conditional (EMA 0.130).}\label{fig:h_sweep}
\end{figure}

\paragraph{Hyperparameters} \cref{fig:h_sweep} depicts a sensitivity analysis of the perturbation size parameter $h$ used in SFG's finite difference computation with EDM2 conditional generation. We observed little sensitivity to $h$, but we chose $h=0.1$ for the remaining experiments due to the slightly improved FD-DINOv2 scores. When feasible, we depict metrics for multiple choices for $\alpha$, SFG's shift parameter. For SEG, we set the attention smoothing parameter $\sigma_A=\infty$ since it achieved the best FID in their experiments \cite{hong2024smoothed}. For comparisons with unguided and auto-guided EDM2 networks, we employ the best hyperparameters reported by \citet{karras2024guiding} for both FID and FD-DINOv2. All models are sampled using 100 steps of the 2nd order Heun solver.

\begin{figure}[ht]
    \centering
    \includegraphics[width=0.95\linewidth,trim=5 0 0 0, clip]{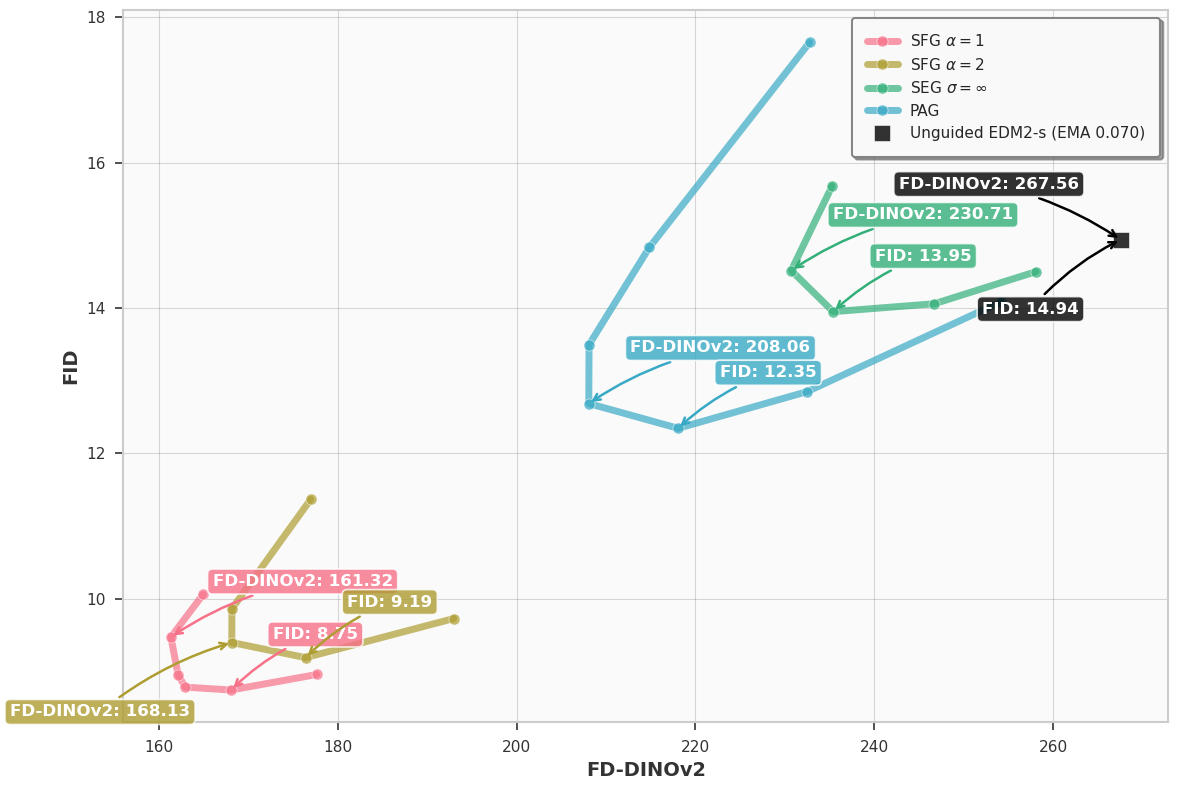}
    \caption{Comparison of FD-DINOv2 and FID scores for single-model unconditional guidance methods with EDM2-s on ImageNet-512.}\label{fig:uncond_comparison}
\end{figure}

\paragraph{Unconditional ImageNet} We compare the performance of the Group (2) guidance methods, SEG, PAG, and SFG, in \cref{fig:uncond_comparison}. Each is created by increasing the guidance weight of each method and measuring FID and FD-DINOv2. The baseline unguided model, EDM2-s-unconditional, yields the highest (worst) metrics with scores of 267 in FD-DINOv2 and 14.94 in FID. Starting from the baseline model, we sweep $w_{\text{SFG}} \in [1, 10]$, $w_{\text{SEG}} \in [1.01, 1.2]$, and $w_{\text{PAG}} \in [1.01, 1.15]$. All unconditional guidance methods improve the FD-DINOv2 and FID scores, with SFG achieving the lowest (best) scores while SEG and PAG yield moderate improvement. SEG decreases DINOv2 by ~14\% and PAG decreases DINOv2 by ~22\%. SFG lowers FD-DINOv2 and FID significantly, with a decrease of ~40\% in each score.

\begin{figure}[ht]
    \centering
    \includegraphics[width=0.95\linewidth,trim=5 0 0 0, clip]{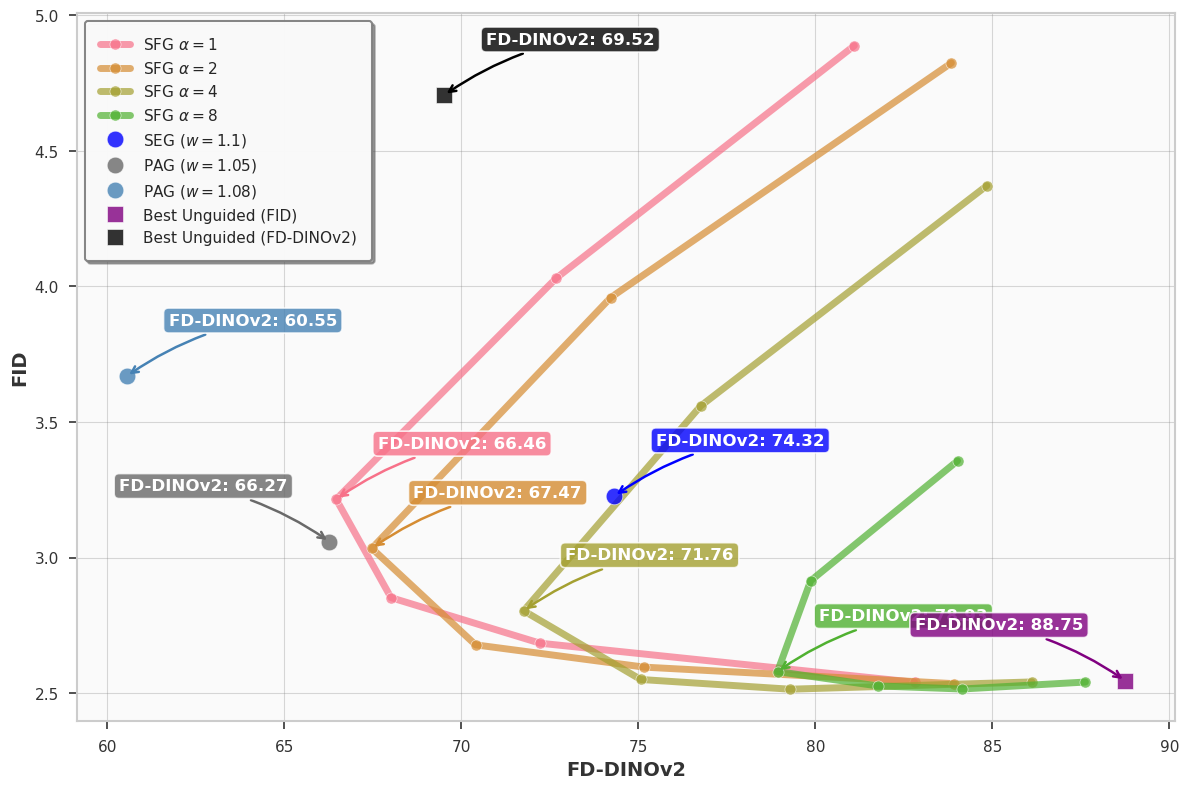}
    \caption{FD-DINOv2 and FID scores for single-model EDM2-s-conditional on ImageNet-512.}\label{fig:cond_comparison}
\end{figure}

\paragraph{Conditional ImageNet} We evaluate SFG on ImageNet-512 generation with EDM2-s-conditional in \cref{fig:cond_comparison}. Starting from the best performing EDM2 model for the FID metric, we apply SFG of various $\alpha$ and sweep $w_{\text{SFG}} \in [1, 30]$, forming the curves. SFG trades off FID to lower the FD-DINOv2 of the baseline model by ~25\% from 88.75 to 66.46, forming a Pareto frontier. PAG appears to follow the same Pareto frontier, but PAG shows more robustness to stronger guidance, enabling further FID to FD-DINOv2 tradeoff than is feasible with SFG. Compared with the best performing ED2-s-conditional model for FD-DINOv2, SFG with $\alpha=1$ lowers the FD-DINOv2 by an additional $3$ points while reducing FID by 20\%.

\paragraph{Combining SFG with Auto-Guidance} Auto-Guidance trains a less capable version of the original model to guide generation, achieving state-of-the-art FID and FD-DINOv2 \cite{karras2024guiding}. We try a simple experiment in which we add SFG to the Auto-Guidance (AG) method, where the output of the AG system is input to SFG as though the AG system was a single model. We evaluate the method for various values of $\alpha$ on ImageNet-512 unconditional generation. The results are depicted in \cref{subfig:ag_sfg_uncond}. Increasing $w_{\text{SFG}}$ trades off a moderate amount of FID for a substantial decrease of ~10\% in FD-DINOv2, achieving a new state-of-the-art in ImageNet-512 unconditional generation. Compared with the best AG-only models for FD-DINOv2, adding SFD lowers FD-DINOv2 by an additional 8 points without increasing FID.

\paragraph{SFG and DiT Unconditional Sampling} We apply SFG to DiT-XL-2-512 \cite{peebles2023scalable} to enhance ImageNet-512 unconditional generation. The original unguided model achieved a FID of $~48.7$, and adding SFG ($\alpha=1$, $w_{\text{SFG}}=5)$ reduced FID by $~25\%$ to $36.85$. We did not try other SFG hyperparameters due to computational constraints. Note that PAG and SEG were designed specifically for the U-Net architecture and therefore could not be directly applied to DiT. 

\subsection{Text-to-Image Experiments}

We implement SFG for the text-to-image flow matching transformers FLUX.1-dev and Stable Diffusion 3.5-medium \cite{esser2024scaling}. Flow estimates $v_\theta(\vx, t)$ are converted to noise estimates via the paramterization $\rveps_\theta(\vx, t) = (1-t)v_\theta(\vx, t) + \vx$ and subsequently input to SFG. SFG guides the noise estimate, and the guided noise estimate $\tilde{\rveps_\theta}(\vx, t)$ is converted back to a flow estimate with $\tilde{v_\theta}(\vx, t) = \left(\tilde{\rveps_\theta}(\vx, t) - \vx\right)\big/(1-t)$.

\begin{figure*}[ht]
    \centering
    \hfill
    \begin{subfigure}{0.49\linewidth}
        \includegraphics[width=\linewidth]{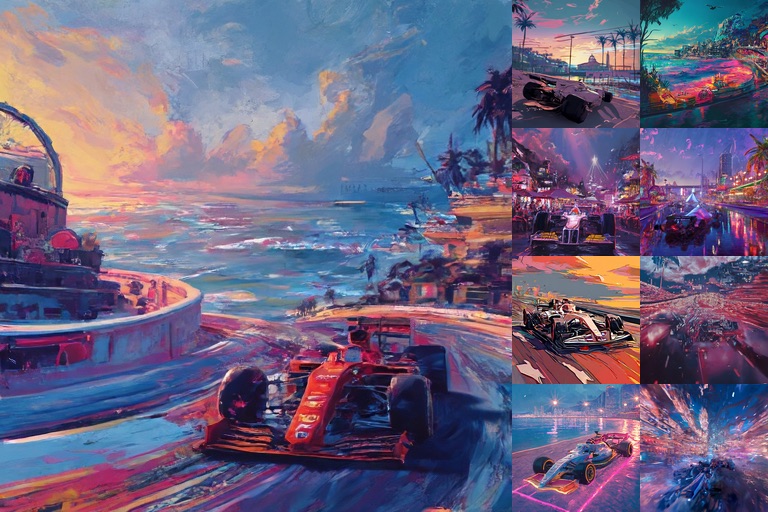}
        \caption{No Guidance: \textcolor{red}{Low Image Fidelity}}
    \end{subfigure}
    \hfill
    \begin{subfigure}{0.49\linewidth}
        \includegraphics[width=\linewidth]{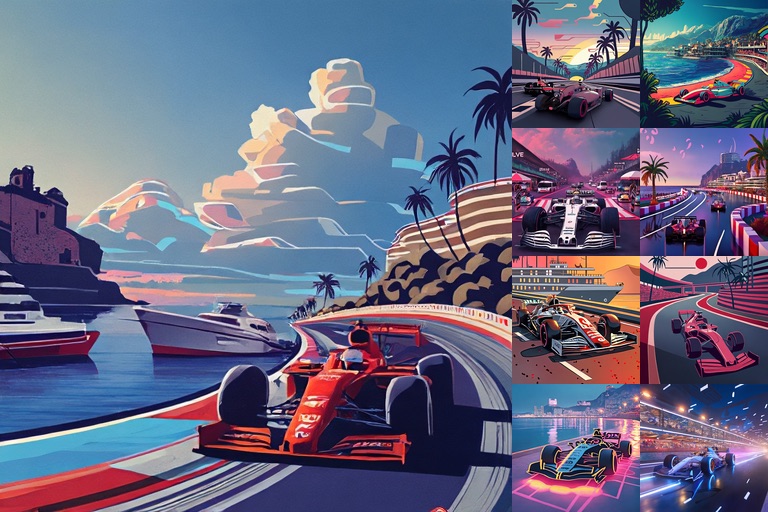}
        \caption{SFG = 20 ($\alpha=1$): \textcolor{blue}{High Fidelity} and \textcolor{blue}{High Diversity}}
    \end{subfigure}
    \hfill

    \hfill
    \begin{subfigure}{0.49\linewidth}
        \includegraphics[width=\linewidth]{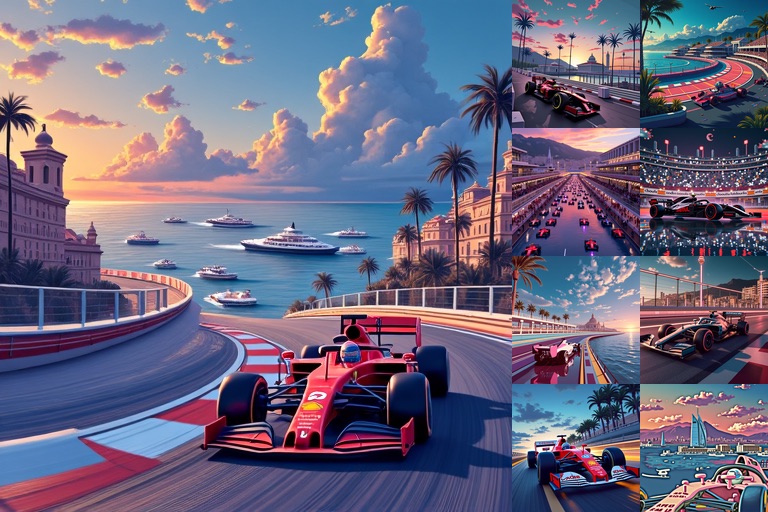}
        \caption{CFG = 7 \& Interval (0.1, 0.8): \textcolor{blue}{High Fidelity} and Moderate Diversity}
    \end{subfigure}
    \hfill
    \begin{subfigure}{0.49\linewidth}
        \includegraphics[width=\linewidth]{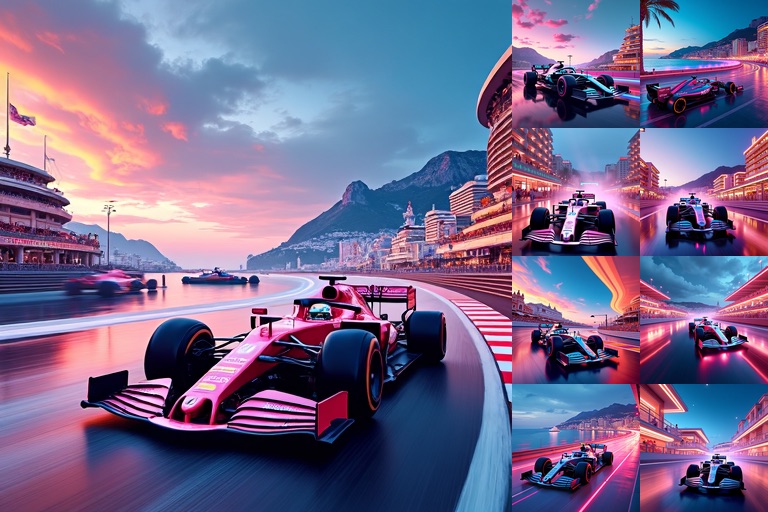}
        \caption{CFG = 3.5: \textcolor{blue}{High Fidelity} and \textcolor{red}{Low Diversity}}
    \end{subfigure}
    \hfill
    \caption{FLUX.1-dev generated images of \textit{``Vaporwave art of the F1 grand prix in Monaco''} using no guidance, Classifier-Free Guidance (CFG), and Saddle-Free Guidance (SFG, ours).}\label{fig:flux_experiment}
\end{figure*}

\paragraph{FLUX.1-dev Comparison} We sample FLUX.1-dev without guidance, with CFG (distilled) of varying weight, and with SFG ($\alpha=1$, $w_{\text{SFG}}=20$) with the prompt \textit{``Vaporwave art of the F1 grand prix in Monaco.''} The resulting images are displayed in \cref{fig:flux_experiment}. Sampling without guidance results in fuzzy images lacking coherent structure. At high levels of CFG the output distribution is very narrow and lacks diversity, depicting highly similar airbrushed scenes. Adding a guidance interval improves diversity of the generated images at the cost of some prompt adherence. Sampling with SFG results in high-fidelity samples which adhere most closely to the prompt, clearly depicting the Monaco Grand Prix in true Vaporwave style. The distribution of images from SFG is diverse, depicting various plausible scenes with a rich color palette.

\begin{figure*}[ht]
    \centering
    \hfill
    \begin{subfigure}{0.49\linewidth}
        \includegraphics[width=\linewidth]{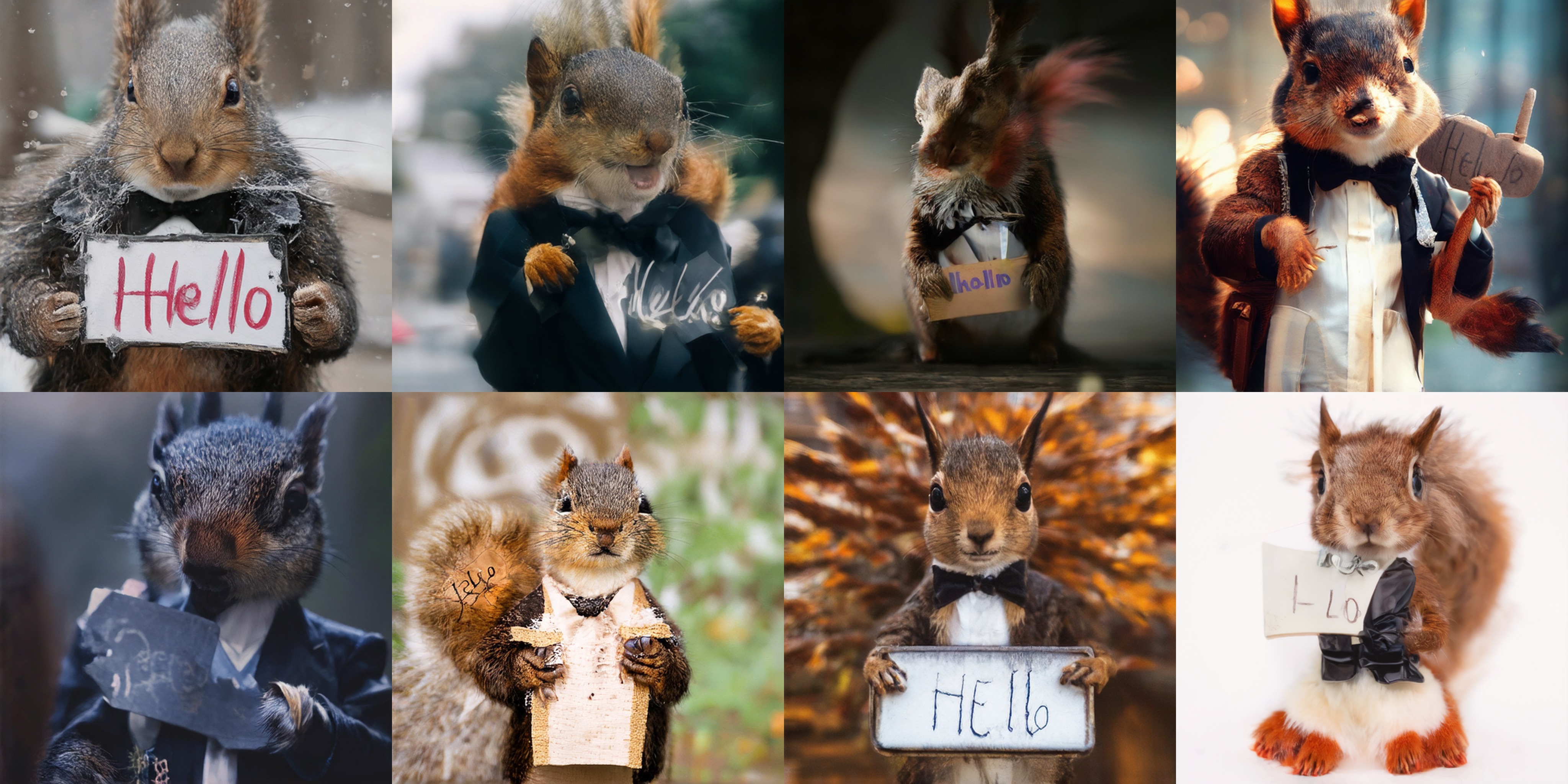}
        \caption{No Guidance: \textcolor{red}{Low Image Fidelity}}
    \end{subfigure}
    \hfill
    \begin{subfigure}{0.49\linewidth}
        \includegraphics[width=\linewidth]{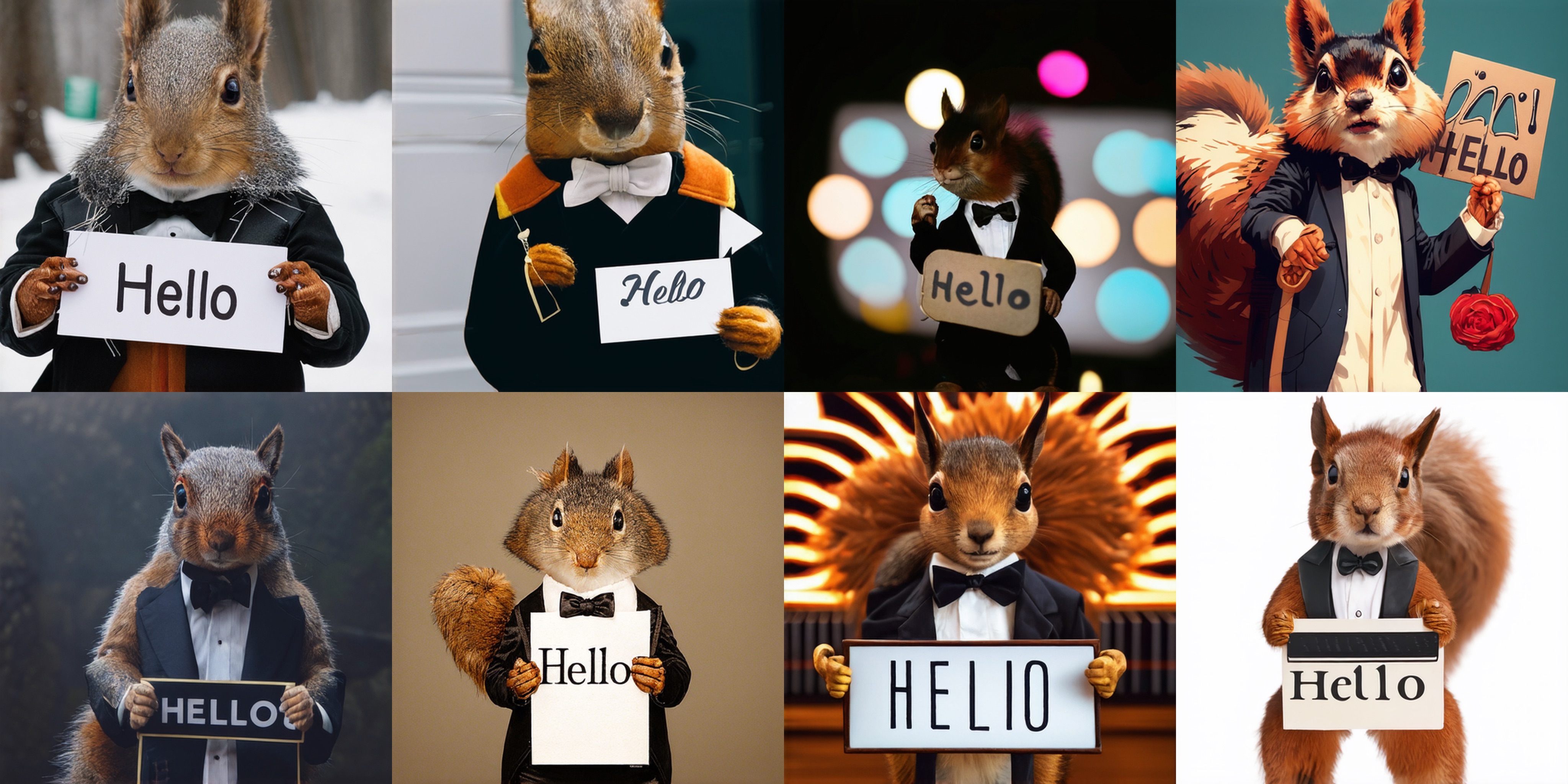}
        \caption{SFG = 20 ($\alpha=1$): \textcolor{blue}{High Fidelity} and \textcolor{blue}{High Diversity}}
    \end{subfigure}
    \hfill

    \hfill
    \begin{subfigure}{0.49\linewidth}
        \includegraphics[width=\linewidth]{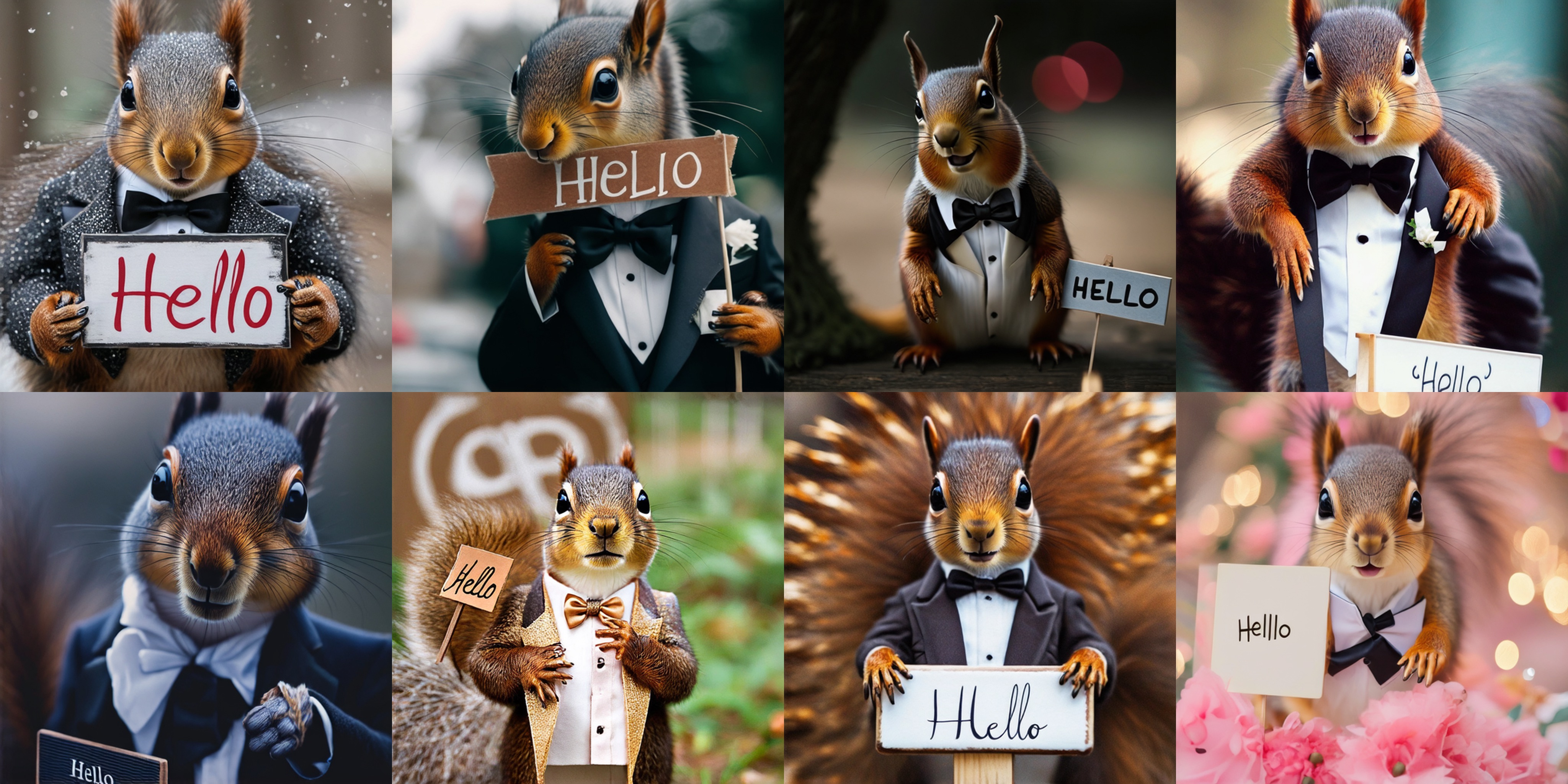}
        \caption{CFG = 5 \& Interval (0.1, 0.8): Moderate Fidelity and Moderate Diversity}
    \end{subfigure}
    \hfill
    \begin{subfigure}{0.49\linewidth}
        \includegraphics[width=\linewidth]{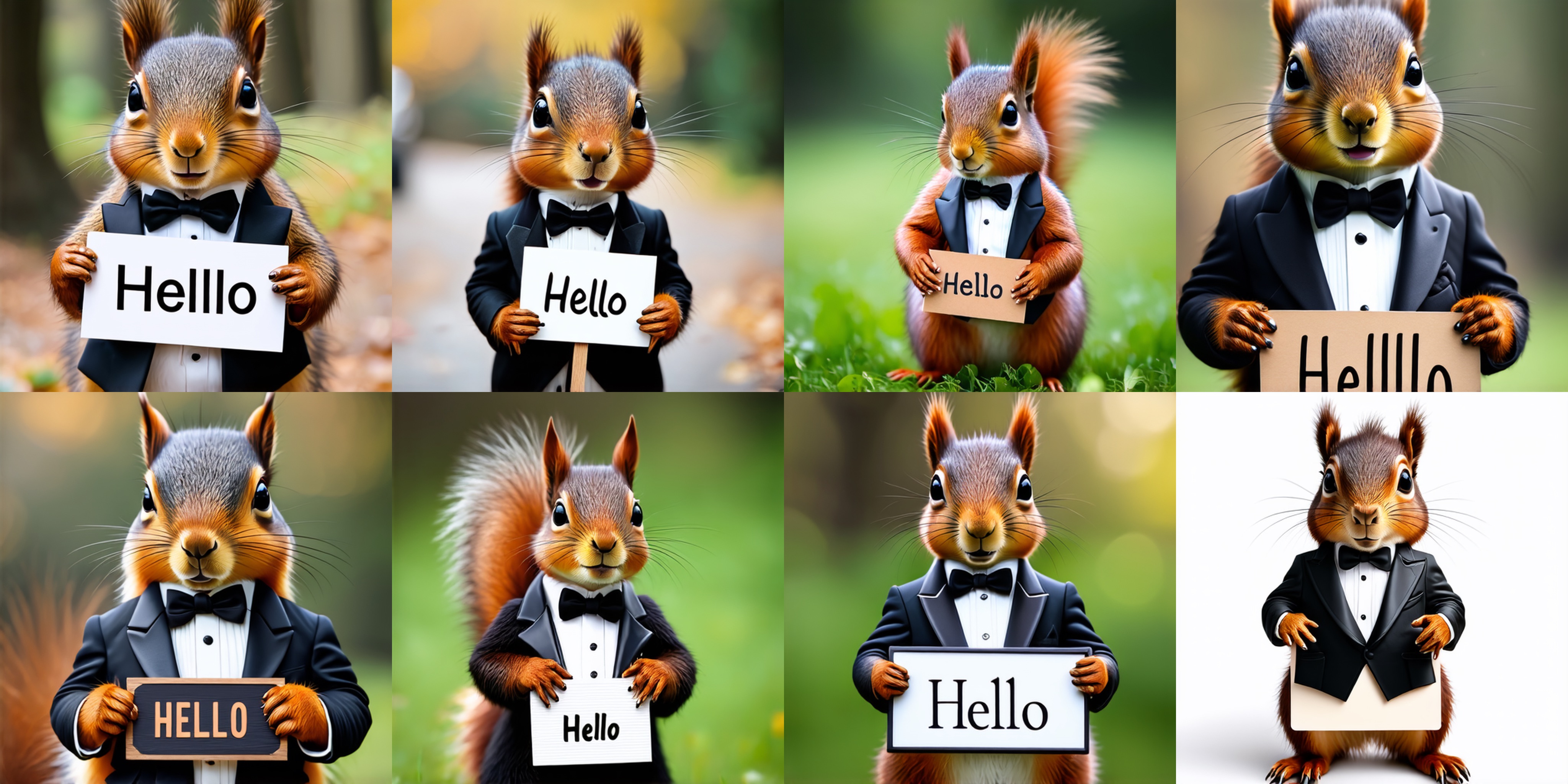}
        \caption{CFG = 5: \textcolor{blue}{High Fidelity} and \textcolor{red}{Low Diversity}}
    \end{subfigure}
    \hfill
    \caption{SD3.5-medium generated images of \textit{``A photo of a squirrel in a tuxedo with a sign that says `Hello'.''} using no guidance, Classifier-Free Guidance (CFG), and Saddle-Free Guidance (SFG, ours).}
\end{figure*}

\paragraph{SD3.5-medium Comparison} We run a similar experiment with Stable Diffusion 3.5-medium (SD3.5) \cite{esser2024scaling} using the prompt \textit{``A photo of a squirrel in a tuxedo with a sign that says `Hello'. ''} Sampling with no guidance results in vague images lacking clear structure, and few of the images adhere closely to the prompt. Sampling with CFG significantly boosts prompt adherence and image fidelity, but image diversity is sharply reduced. All subjects of the images share a similar pose and airbrushed `AI generated' appearance. Adding an interval sacrifices some image fidelity to promote image diversity, however the subjects still exhibit the airbrushed appearance. Sampling with SFG yields diverse, coherent images which adhere well to the prompt. Subjects have varying poses, appearances, color schemes, and levels of photorealism.
\section{Discussion}

\paragraph{Limitations} Our analysis of guidance methods is limited to the image generation setting and we focus heavily on two metrics: FID and FD-DINOv2. There are many other interesting applications of diffusion and flow matching models that would make informative benchmarks for guidance methods. We do not perform an study the effect of power iteration count on SFG sample quality - we fix the parameter to 1 iteration per sampling step in all experiments. We do not include extensive experiments with DiT \cite{peebles2023scalable} due to computational constraints and the relative lack of inference-time unconditional guidance methods which are compatible with it.

\paragraph{Societal Impacts and Ethical Consideration} Our work concerns a new guidance strategy for deep generative models. Generative models learn to mimic their training data, and therefore our method could produce samples which perpetuate existing biases from the data. Moreover, our method enhances the realism of generated data, potentially making the generative models more useful for bad actors. Evaluating generative models is a computationally costly procedure, and this work potentially led to increased emissions.

\section{Conclusion}

Current guidance methods rely on differences between a primary score model and an entropic reference model to correct generation. The dependence on a reference model limits the applicability of guidance methods, as it necessitates training additional models, having labeled data, or intervening on model-specific computation. We challenge the reference based guidance paradigm and make the surprising discovery that the positive curvature of the saddle regions of the primary score model provides a strong guidance signal.

We introduce Saddle-Free Guidance (SFG), a training-free and model-agnostic guidance method which steers sampling with the maximal positive curvature of the learned log density. We compare SFG with other guidance methods on the ImageNet-512 generation task and show that it achieves state-of-the-art single-model unconditional generation in terms of FID and FD-DINOv2 scores. When SFG is combined with Auto-Guidance, it reduces FD-DINOv2 by an additional 10\% to achieve a general state-of-the-art in unconditional ImageNet-512 generation. We implement SFG for FLUX.1-dev and Stable Diffusion 3.5-medium and show that SFG promotes image fidelity while maintaining strong prompt adherence and sample diversity.
{
    \small
    \bibliographystyle{ieeenat_fullname}
    \bibliography{main}
}

\clearpage
\setcounter{page}{1}
\maketitlesupplementary

\section{Additional Reproducibility Information}

\subsection*{Simplex Experiment}

DiT models are initialized with the architecture configurations in \cref{tab:dit_arches}. The training dataset consists of 1000 samples from the GMM, and the DiT models are trained for 30000 batches of size 200 with a denoising score matching objective. The training process uses 500 warm up batches, an initial learning rate of $1e-3$ with a cosine annealing schedule and the Adam optimizer with a weight decay of $1e-5$.

The models are tested with 1000 samples drawn from the mode, outlier, and saddle distributions described in the main text. Losses at ``flow matching times'' are computed by scaling the samples and adding noise according to the forward flow matching process.

\begin{table}[h]
  \caption{Diffusion Transformer (DiT) Architecture Configurations for the Simplex Experiment.}
  \label{tab:dit_arches}
  \centering
  \begin{tabular}{@{}lcccc@{}}
    \toprule
    Model & Layers & Patch Size & Encoding Dim. & Heads \\
    \midrule
    DiT-S & 2 & 4 & 16 & 4 \\
    DiT-M & 6 & 4 & 64 & 8 \\
    DiT-L & 10 & 4 & 256 & 16 \\
    \bottomrule
  \end{tabular}
\end{table}

\subsection*{EDM2 ImageNet512 Experiments}

All score models (all EDM2, DiT-XL-2-512, FLUX.1-dev, SD3.5-m) are sampled with 100 steps in every experiment. EDM2 uses the EDM sampler \cite{karras2022elucidating}, DiT uses the DDIM sampler \cite{song2020denoising}, and FLUX.1-dev and SD3.5-m use the Euler sampler.

\paragraph{Unconditional ImageNet} We employ EDM2-s-0.070 \cite{karras2024analyzing} in the single model unconditional generation experiments, where 0.070 denotes the EMA length. For Auto-Guidance we use EDM2-xs-0.110 (trained on 1/16th the number of images) as the guiding model since that combination led to lowest FID in \cite{karras2024guiding}. 

\paragraph{Conditional ImageNet} We employ EDM2-s-0.130 in single model conditional generation experiments, where 0.130 refers to the EMA length. This is the best reported EDM2 for single model ImageNet-512 conditional FID.

\section{Gaussian Separation Problem}

\begin{figure}[ht]
    \centering
    \includegraphics[width=0.99\linewidth,trim=10 30 10 30,clip]{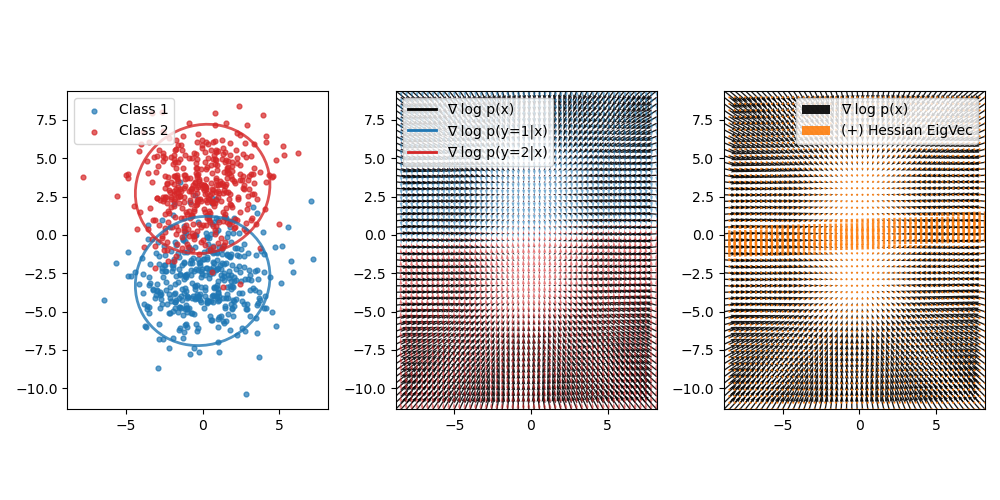}
    \caption{Additive variance $\sigma^2=4$}
    \label{fig:gaussian_sep_4}
\end{figure}

\begin{figure}[ht]
    \centering
    \includegraphics[width=0.99\linewidth,trim=10 30 10 30,clip]{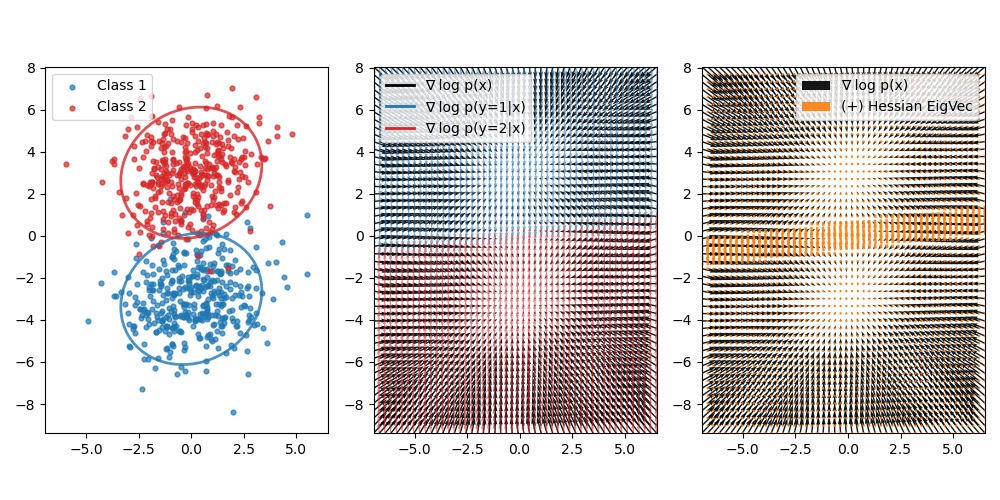}
    \caption{Additive variance $\sigma^2=2$}
    \label{fig:gaussian_sep_2}
\end{figure}

\begin{figure}[ht]
    \centering
    \includegraphics[width=0.99\linewidth,trim=10 30 10 30,clip]{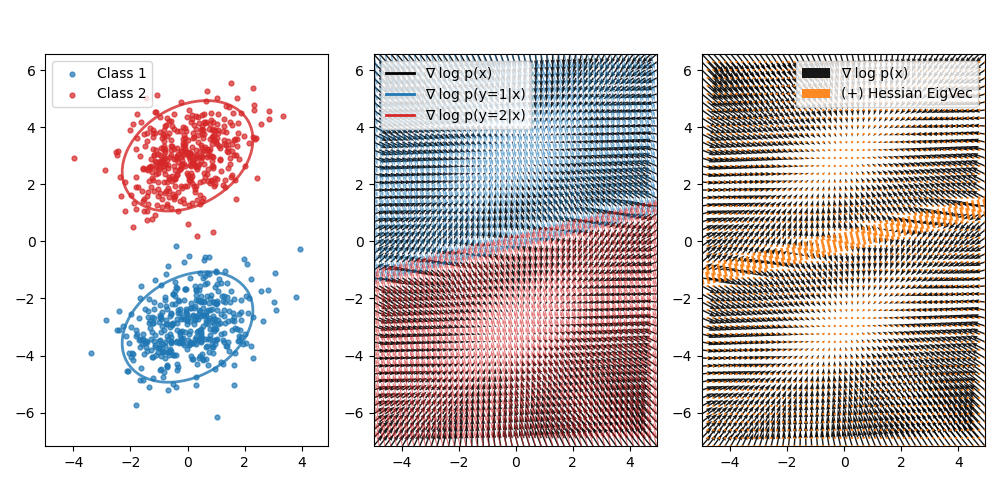}
    \caption{Additive variance $\sigma^2=0.5$}
    \label{fig:gaussian_sep_0_5}
\end{figure}

We construct a toy classification problem consisting of two identical Gaussians in 2D. Different amounts of isotropic Gaussian noise are added to the two components to mimic a diffusion model generation scenario, and the marginal score $\nabla \log p(\vx)$ as well as the ``classifier'' gradients (e.g., $\nabla \log p(y=1|\vx)$) are plotted. Lastly, we plot the eigenvectors corresponding to positive curvature on top of the marginal score. See \cref{fig:gaussian_sep_4,fig:gaussian_sep_2,fig:gaussian_sep_0_5} for the results.

The plots show that the marginal score experiences a saddle point between the two classes. Classifier gradients $\nabla \log p(y=0|\vx)$ and $\nabla \log p(y=1|\vx)$ move data away from outlier and saddle regions and toward modes. Similarly, positive curvature can be leveraged to move data outside of saddle regions in a manner similar to classifier free guidance.

\begin{figure*}[ht]
    \centering
    \hfill
    \begin{subfigure}{0.49\linewidth}
        \includegraphics[width=\linewidth]{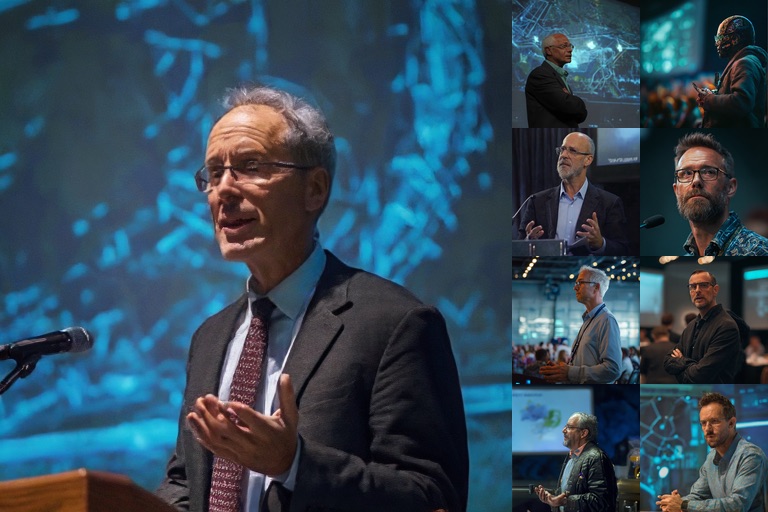}
        \caption{No Guidance: \textcolor{red}{Low Image Fidelity}}
    \end{subfigure}
    \hfill
    \begin{subfigure}{0.49\linewidth}
        \includegraphics[width=\linewidth]{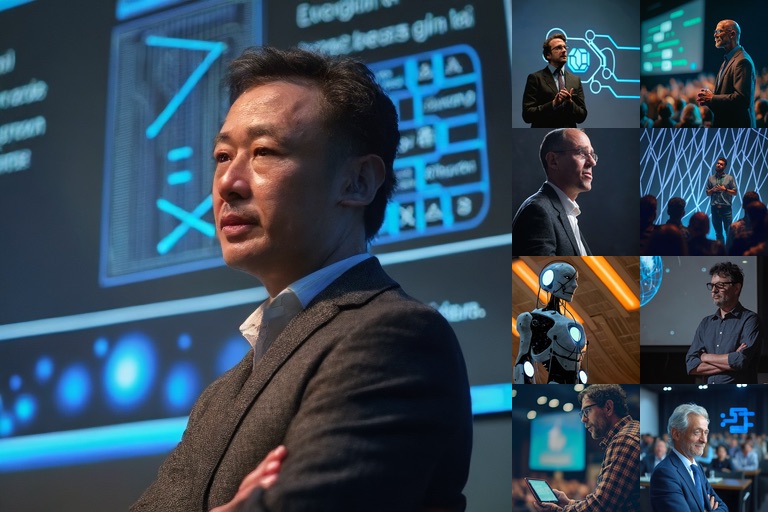}
        \caption{SFG = 20 ($\alpha=1$): \textcolor{blue}{High Fidelity} and \textcolor{blue}{High Variation}}
    \end{subfigure}
    \hfill

    \hfill
    \begin{subfigure}{0.49\linewidth}
        \includegraphics[width=\linewidth]{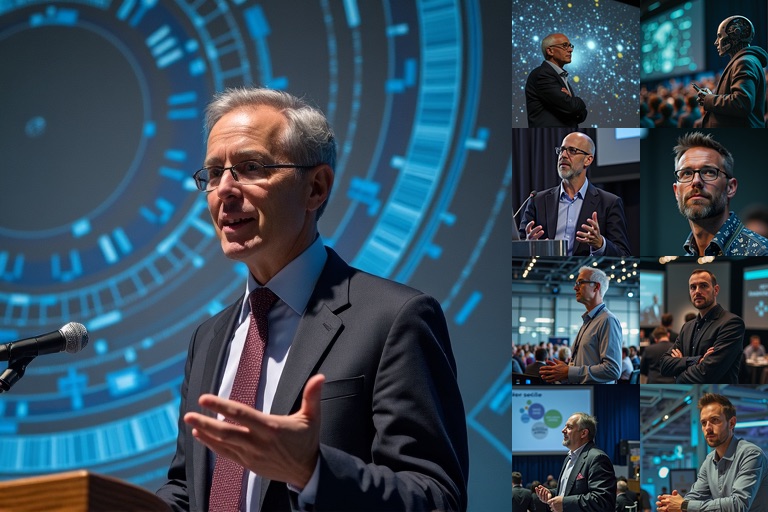}
        \caption{CFG = 7 + Interval (0.1, 0.8): \textcolor{blue}{High Fidelity} and Moderate Variation}
    \end{subfigure}
    \hfill
    \begin{subfigure}{0.49\linewidth}
        \includegraphics[width=\linewidth]{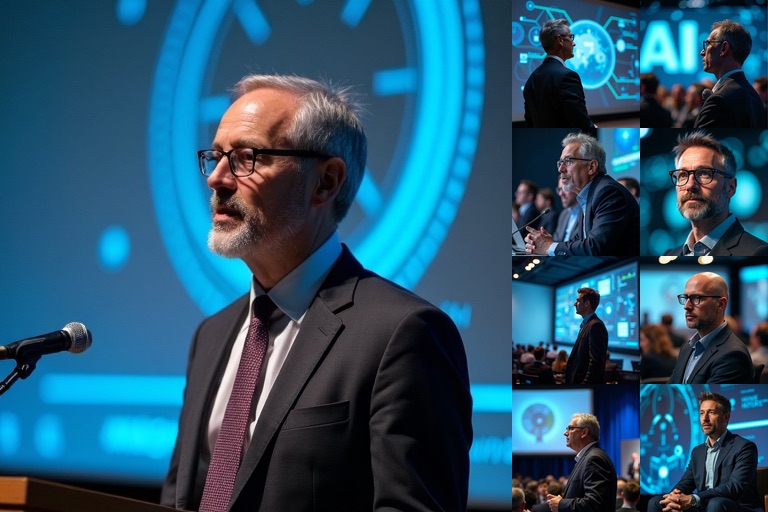}
        \caption{CFG = 3.5: \textcolor{blue}{High Fidelity} and \textcolor{red}{Low Variation}}
    \end{subfigure}
    \hfill
    \caption{FLUX.1-dev generated images of \textit{``A photograph of a scientist at a conference about AI.''} using no guidance, Classifier-Free Guidance (CFG), CFG in the Interval (0.1, 0.8), and Saddle-Free Guidance (SFG, ours).}\label{fig:flux_experiment_ai}
\end{figure*}

\end{document}